\documentclass[manuscript,screen,nonacm]{acmart}

\usepackage{graphics}
\usepackage{hyperref}
\usepackage{algorithm}
\usepackage[algo2e, ruled, lined, vlined, longend]{algorithm2e}

\graphicspath{{figures}}

\AtBeginDocument{%
  }

\acmConference{}
\renewcommand\footnotetextcopyrightpermission[1]{} % removes footnote with conference info
\setcopyright{none}

\begin{document}
% \title{NSGA-II with Facilitated Variation Applied to the Extended Oxley Model for Orthogonal Metal Cutting}
% Sebastian: I consulted Alex. He suggests that "Metal Cutting Process Optimization" is a better choice than "CAM parameter optimization"
%\title{Dynamic NSGA-II for Multi-Objective Metal Cutting Process Optimization}
\title{Studying Evolutionary Solution Adaption Using a Flexibility Benchmark Based on a Metal Cutting Process}
\author{L\'{e}o Fran\c{c}oso Dal Piccol Sotto}
% \email{leo.sotto@scai.fraunhofer.de}
\orcid{0000-0002-6406-7322}
\author{Sebastian Mayer}
% \email{sebastian.mayer@scai.fraunhofer.de}
\affiliation{
  \institution{Fraunhofer SCAI}
  \city{Sankt Augustin}
  \country{Germany}
}
\author{Hemanth Janarthanam}
\email{hemanth.janarthanam@iwm.fraunhofer.de}
\author{Alexander Butz}
\email{alexander.butz@iwm.fraunhofer.de}
\orcid{0009-0006-7930-1237}
\affiliation{
  \institution{Fraunhofer IWM}
  \city{Freiburg}
  \country{Germany}
}
\author{Jochen Garcke}
\email{jochengarcke@scai.fraunhofer.de}
\orcid{0000-0002-8334-3695)}
\affiliation{
  \institution{Fraunhofer SCAI}
  \city{Sankt Augustin}
  \country{Germany}
}
\affiliation{%
  \institution{University of Bonn}
  \city{Bonn}
  \country{Germany}
}


\begin{abstract}
We consider optimizing for different production requirements from the viewpoint of a bio-inspired framework for system flexibility that allows us to study the ability of an algorithm to transfer solutions from previous optimization tasks, which also relates to dynamic evolutionary optimization.

Optimizing manufacturing process parameters is typically a multi-objective problem with often contradictory objectives such as production quality and production time. If production requirements change, process parameters have to be optimized again. Since optimization usually requires costly simulations based on, for example, the Finite Element method, it is of great interest to have means to reduce the number of evaluations needed for optimization. Based on the extended Oxley model for orthogonal metal cutting, we introduce a multi-objective optimization benchmark where different materials define related optimization tasks.

We use the benchmark to study the flexibility of NSGA-II, which we extend by two variants: 1) varying goals, which optimizes solutions for two tasks simultaneously to obtain in-between source solutions expected to be more adaptable, and 2) active-inactive genotype, which accommodates different possibilities that can be activated or deactivated. Results show that adaption, i.e. transferring a solution from a previous optimization task, with standard NSGA-II greatly reduces the number of evaluations required for optimization to a target goal in comparison to starting from scratch. The proposed variants further improve the adaption costs, although further work is needed towards making the methods advantageous for real applications.
\end{abstract}

%\begin{CCSXML}
%<ccs2012>
%  <concept_id>10010520.10010553.10010562</concept_id>
%  <concept_desc>Computer systems organization~Embedded systems</concept_desc>
%  <concept_significance>500</concept_significance>
%  <concept_id>10010520.10010575.10010755</concept_id>
%  <concept_desc>Computer systems organization~Redundancy</concept_desc>
%  <concept_significance>300</concept_significance>
%  <concept_id>10010520.10010553.10010554</concept_id>
%  <concept_desc>Computer systems organization~Robotics</concept_desc>
%  <concept_significance>100</concept_significance>
%  <concept_id>10003033.10003083.10003095</concept_id>
%  <concept_desc>Networks~Network reliability</concept_desc>
%  <concept_significance>100</concept_significance>
%</ccs2012>
%\end{CCSXML}

%\ccsdesc[500]{Computer systems organization~Embedded systems}



\maketitle

\section{Introduction}
\label{sec:introduction}

A manufacturing process is a series of steps that transforms raw materials, components, or parts into finished products that meet specific requirements or specifications. Achieving optimal operations typically requires a multi-objective optimization of process parameters in terms of various aspects affecting material cost, product quality, and production time. Evolutionary multi-objective optimization of manufacturing process parameters has received considerable attention~\cite{pereira2021review}. However, these works assume fixed production requirements. If the production requirements change, one has to compute optimal parameters from scratch again. This can be prohibitive if requirements change frequently, in particular, if costly numeric simulations are involved. For instance, simulations based on the Finite Element (FE) method can easily take hours to even days for one simulation run~\cite{pfrommer2018optimisation}. Hence, it is of interest to have flexible evolutionary optimization methods that efficiently adapt solutions from previous optimization tasks to a new related optimization task using only few additional evaluations of the objective functions.

To implement a flexibility benchmark that allows us to measure an evolutionary algorithm's capacity to adapt solutions from previous to new related optimization tasks, we follow the approach used in~\cite{sotto2022pole}, which is based on a general framework for system flexibility introduced in~\cite{mayer-2023}. 
The manufacturing process that we consider in this paper is orthogonal metal cutting. Cutting is a process in which a material is cut to a desired final shape and size by a controlled material-removal process. 
We consider a so-called task context, which is a set of related parameter optimization tasks for the orthogonal cutting process that arises from considering different materials as the production requirement changes. In other words, different materials constitute different but related optimization tasks. As a performance indicator for successful solution adaption we evaluate the overall cost that an evolutionary optimization algorithm has for adapting solutions from a source to a target task. To simulate the cutting process, we use the extended Oxley model~\cite{oatao25920}. This is an analytic model that is inexpensive to evaluate. We chose to use this model as a substitute for costly but more accurate FE simulations to provide a benchmark that allows for extensive experimentation with high repetition numbers that guarantee stable results. Moreover, an easy to evaluate benchmark makes it easier to test many different optimization methods. The downside is that without further validation using realistic simulations, the provided benchmark allows only for negative conclusions which are nevertheless useful. Optimization methods that already need many evaluations to adapt process parameters for the Oxley model do not even have to be considered in more realistic setups. Optimization methods that show good adaption capabilities in the benchmark only yield a first proof-of-concept that requires further validation with more realistic simulations. This validation is beyond the scope of the paper. 

The flexibility setup described above shares similarities with dynamic (multi-objective) optimization~\cite{branke2012evolutionary, azzouz2017dynamic}, which studies optimization problems that change over time. Indeed, one can consider the material changes as discrete events in time that change aspects of the underlying optimization problem. However, in our setup there is no need to detect the changes algorithmically, which is an important aspect of dynamic optimization. The optimization problems usually considered in dynamic optimization are intrinsically dynamic such as the moving peaks benchmark, the dynamic knapsack problem or the dynamic travelling salesman problem \cite{yang2015dynamic}. We are aware of only one work that overlaps with simulation-based multi-objective optimization for manufacturing, in which the turning of material with continuously changing properties, such as gradient materials, is studied~\cite{ROY2008429}. Nevertheless, adaption and transfer of solution is an important aspect of dynamic (multi-objective) optimization that has been intensively studied and which we can build upon. 

Using the new flexibility benchmark, this paper provides an exemplary investigation of the potential of dynamic variants of the well-known NSGA-II algorithm~\cite{Deb2011} for the adaption of manufacturing process parameters. We study two variants: optimizing for two tasks at the same time (varying goals), and using a genotype with active and inactive positions that can accommodate different solutions in one chromosome (active-inactive genotype). Furthermore, instead of adapting a solution to solve each problem in an optimal or near-optimal way as the goals change in time, we aim at providing a good starting point for adaption that is not necessarily optimal for the problems it was trained on. Our results show that: 1) For the defined problem, the cost for adapting solutions from source materials is much lower than the cost for searching from scratch for each material, and 2) The proposed variants are able to further reduce this cost by optimizing source solutions that are in-between solutions for different tasks and that accommodate different values for the process parameters that can be activated or deactivated. However, to be really interesting for industrial use-cases, one should achieve evaluation numbers needed for adaption that are a magnitude lower than what we could achieve in our experiments even in the best. Hence, there is clearly a need for further research.  

The rest of this paper is organized as follows. Section \ref{sec:background} presents some background concepts important for this work: the notion of system flexibility (section \ref{sec:system-flexibility}), the extended Oxley model (section \ref{sec:oxley-model}), and the NSGA-II algorithm (section \ref{sec:nsga-ii}). Section \ref{sec:methodology} presents our proposed benchmark using the extended Oxley model. How we integrate some ideas from facilitated variation into NSGA-II is given in section \ref{sec:nsga-ii-facilitated-variation}. We then present the experimental setup for evaluating our proposed methodology and the results obtained in section \ref{sec:results}. Section \ref{sec:conclusion} presents some concluding remarks and possibilities of future work.

\section{Background}
\label{sec:background}

\subsection{System Flexibility}
\label{sec:system-flexibility}

System flexibility refers to a system's ability (i) to easily adapt from being good at one task to being good at a related task, and (ii) to cope with a diversity of related tasks. The paper~\cite{mayer-2023} introduces a general formalism for system flexibility that allows to define both aspects (i) and (ii) rigorously. In this paper, we focus on aspect (i). 
In preparation for the benchmark defined in section~\ref{sec:definition-task}, we introduce some terminology and cost notions in this section. Thereby, we follow~\cite{sotto2022pole} which has studied evolutionary algorithm from the viewpoint of system flexibility for the pole balancing problem.

We denote the system configuration space by~$X$ and a concrete system configuration by $x \in X$. 
As our application example we consider an orthogonal cutting process and the problem of adapting process parameters when a material change occurs.
A cutting machine with the cutting tool forms a system that can perform the cutting process. The input of the cutting process is a workpiece of some metal and the output is the workpiece with the desired amount of material removed. 
Here, the system configuration space is given by the vector space of all possible values for the process parameters. 

We further consider the system to be equipped with an evolutionary algorithm $A$ that it uses to optimize process parameters. Since the evolutionary algorithm $A$ and how it operates on the system configuration space $X$ are the main concerns in this paper, we formally denote the system by a tuple $M=(A, X)$. In our example, the task $T$ that the system has to perform, for a specific type of metal, is to optimally cut the workpiece according to some feasibility criteria and multiple objectives. For our purposes, we can consider this as equivalent to $A$ solving the associated multi-objective optimization problem. The cost in terms of objective function evaluations to solve this problem from scratch is denoted by $c_0(A,T)$.

To measure the ability of the evolutionary algorithm $A$ to adapt process parameters when a material change occurs, we consider a \textbf{task context} $\mathcal T = \{T_1,\dots,T_n\}$ of $n$ multi-objective optimization problems, each associated to a different type of metal. For each pair $(T_i,T_j)$, where $T_i \ne T_j$, the \textbf{adaption cost} $c_{\textrm{ada}}(A,T_i,T_j)$ denote the cost in terms of objective function evaluations to solve $T_j$ given that the algorithm has previously solved $T_i$. To measure the adaption capability of $A$ with regard to the task context, we can consider the \textbf{worst-case adaption cost} given by
\begin{equation}
\label{eq:worst-case-adaption-cost}
c_{\textrm{ada}}^{\textrm{wor}}(A, \mathcal T) = \max_{T_1 \neq T_2} c_{\textrm{ada}}(A,T_1, T_2),
\end{equation}
the \textbf{average-case adaption cost} given by
\begin{equation}
c_{\textrm{ada}}^{\textrm{avg}}(A, \mathcal T) = \frac{1}{n(n-1)}\sum_{T_1 \neq T_2} c_{\textrm{ada}}(A,T_1, T_2),
\end{equation}
or the \textbf{best-case adaption cost} given by
\begin{equation}
\label{eq:best-case-adaption-cost}
c_{\textrm{ada}}^{\textrm{wor}}(A, \mathcal T) = \min_{T_1 \neq T_2} c_{\textrm{ada}}(A,T_1, T_2),
\end{equation}

The lower the adaption cost, the better is the considered algorithm in exploiting information gained on a previous optimization task for a new optimization task. As reference values, we also consider the cost to solve the optimization task from scratch in the worst-, average-, or best-case.

Note that the above definitions are not specific to evolutionary algorithms but can be used for any optimization algorithms. We only have to provide measures for $c_0$ and $c_{\textrm{ada}}$ that are suitable for the considered optimization algorithm. The specific cost measures considered in this paper are discussed in in section~\ref{sec:results}.

\subsection{Orthogonal Metal Cutting and the Extended Oxley Model}
\label{sec:oxley-model}

Orthogonal metal cutting is a machining operation where the cutting edge of the tool is perpendicular to the direction of relative motion between the tool and the workpiece surface. The process involves the removal of material from the workpiece by the cutting tool, in a series of small, discrete steps, as the tool chips away the workpiece material through plastic deformation. The process is known for producing high-quality, precise cuts, but is also challenging to optimize due to the complex interactions between the cutting tool, the workpiece material, and the machining environment.

Predictive models are extensively developed and used in the process planning phase in order to enhance the product quality and to optimize the process parameters with respect to tool life, surface finish, part accuracy and beyond. Predictive models are divided into analytical models, which describe an idealized underlying physics; empirical models, which are derived from experimental observations; and numerical models, such as finite element methods (FEM), which take into account the precise multiphysical phenomena involved in the cutting process~\cite{Arrazola.2013}. Due to the high costs involved in experimentally determining the empirical
models and high computation power required for numerical methods, they are seldom used in an industrial context. On the contrary, analytical models including \cite{Palmer1959623} are fast and assists in developing practical tools for the industry, albeit with the drawback of not capturing the multidimensional physics. The recent advancements in analytical models have however enhanced the predictions to be more realistic. Due to its simplicity and negligible numerical costs, the analytical model initially proposed by \cite{Palmer1959623} and the extension from \cite{oatao25920} is used in this study. The work of \cite{oatao25920} was implemented as a python package and made available on GitHub\footnote{\url{https://github.com/pantale/OxleyPython}} by the authors \cite{pantale-2022}.

\begin{figure}
\center
\includegraphics[width=0.4\columnwidth]{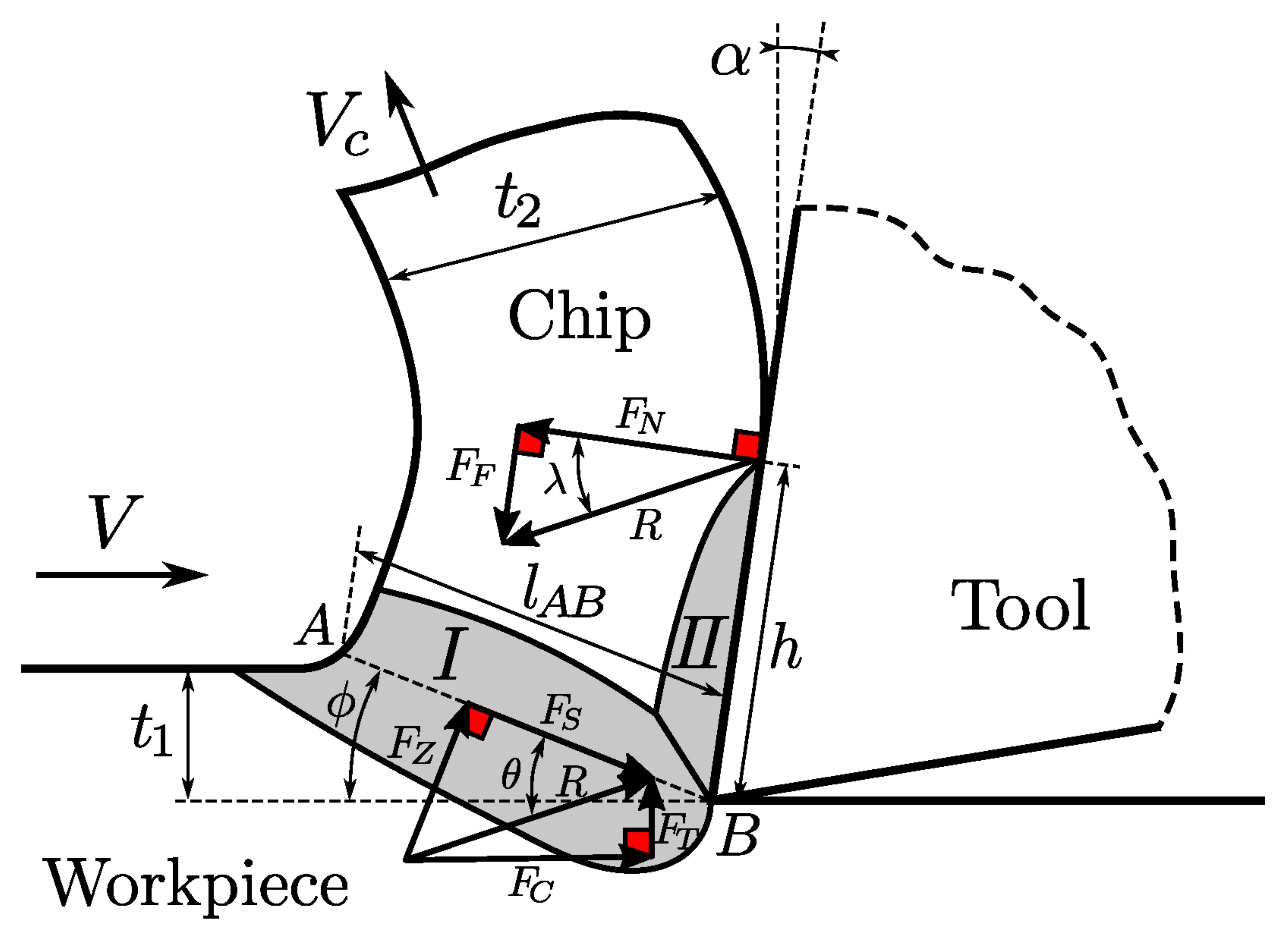}
\caption{A graphical representation of the orthogonal machining model. The image has been taken from \cite{pantale-2022}.}
\label{fig:Oxleymodel}
\end{figure}

Oxley's theory exploits the slip line field theory coupled with thermal phenomena to predict the cutting forces, temperatures and the stresses and strains in the workpiece. The flow stress $\sigma_{y}$ in the workpiece that depends on the magnitude of plastic deformation $\varepsilon_{p}$, rate of plastic deformation $\dot{\varepsilon}_{p}$ and current temperature $T_{w}$ of the material plays a central role in the prediction model. To account for a wide range of materials, a Johnson-Cook material flow rule (Equation \ref{eq:JCook}) that  multiplicatively accounts for each of the influencing phenomena is used. A material is then fully defined by the material parameters plastic hardening parameters $A$, $B$, $n$, plastic deformation rate sensitive parameters $C$, $\dot{\varepsilon}_{0}$, and thermal softening exponent $m$ and the material's melting point $T_{m}$

\begin{equation}
\sigma_{y}=(A+B\varepsilon_{p}^{n})\Big[1+C\ln{\dfrac{\dot{\varepsilon}_{p}}{\dot{\varepsilon}_{0}}}\Big]\Big[1-\Big(\dfrac{T-T_{w}}{T_{m}-T_{w}}\Big)^{m}\Big]
\label{eq:JCook}
\end{equation}

Figure \ref{fig:Oxleymodel} describes the analytical orthogonal cutting model. The material in the vicinity of the tool tip is divided into a primary shear zone (I) of length $l_{AB}$ where the material experiences compressive forces and initiates the plastic deformation along the line AB leading to a chip formation, and a secondary shear zone (II) where a further plastic deformation is induced due to the friction between the chip and tool contact. A workpiece is fed with a velocity of $V$ against the tool to remove a layer of thickness $t_{1}$ resulting in a chip of thickness $t_{2}$ with a velocity of $V_{c}<V$. The primary task of Oxley's theory is to identify three internal variables that depend on the shear angle $\phi$, the ratio of $l_{AB}$ to the thickness primary shear zone and the ratio of chip thickness $t_{2}$ to the thickness of secondary zone, by solving a system of 3 non linear equations. The cutting force $F_{c}$, the advancing force $F_{t}$ and the rise in temperature in the individual zones are then computed from the internal variables. The readers are referred to \cite{pantale-2022} for a detailed description of the algorithm. We stress once more that due to simplicity of the model, one clearly has to expect a gap between the extended Oxley model and realistic simulations. Nevertheless, a benchmark based on the extended Oxley model can provide valuable insights for adapting process parameters in manufacturing contexts as optimization methods that already need many evaluation methods to adapt parameters for the Oxley do not even need to be considered in more realistic settings.

The benchmark in section~\ref{sec:definition-task} is an implementation of the formalism for the orthogonal cutting process and the problem of adapting process parameters when a material change occurs based on the extended Oxley model. 
In the formulation of section~\ref{sec:system-flexibility}, the system configuration space $X$ is given by the vector space of all possible values for the process parameters tool speed $V$, tool rake angle $\alpha$, and the cutting depth in one step $t_1$, see \ref{fig:Oxleymodel}. 

\subsection{NSGA-II: Non-Dominated Sorting Genetic Algorithm II}
\label{sec:nsga-ii}

Many real-world problems require the optimization of two or more objectives at the same time, resulting in the class of multi-objective optimization problems \cite{Deb2011, emmerich2018tutorial}. Different objectives are often contradictory. Therefore, instead of searching for one solution that optimizes all objectives, multi-objective optimization methods search for \textit{a set of non-dominated solutions, the Pareto front}. 

\begin{definition}
\label{def:pareto-front}
Given a set of solutions $X$, a \textbf{Pareto front} $P$ is the set of non-dominated solutions from $X$. A solution $x_i \in X$ is non-dominated if and only if, $\forall x_j \in X$, with $x_j \ne x_i$, $x_i$ is better than $x_j$ in at least one objective.
\end{definition}

The Pareto front is thus the set of trade-off solutions whose objectives cannot be improved without negatively impacting one or more of the other objectives. Given an optimized Pareto front, the decision making for which of the solutions to use will depend on each specific application.

As evolutionary algorithms work with populations of solutions, they are often employed for this class of problems \cite{Deb2011, emmerich2018tutorial}. Among them, the Non-Dominated Sorting Genetic Algorithm (NSGA-II), proposed by \citep{deb2002nsgaii}, is a popular multi-objective optimization algorithm. The NSGA-II algorithm has an overall functionality similar to that of a standard genetic algorithm. An initial population of candidate solutions (individuals) is randomly initialized. Individual chromosomes are represented either as a vector of integers or of floats. Each generation, individuals are evaluated and attributed a fitness score. Tournaments are performed to select individuals based on this score, and these selected individuals are subject to the crossover and mutation operators, until a new population is formed. The search continues for a maximum number of generations or until a desired solution is found. The difference in NSGA-II lies mainly on the selection mechanism, which is supported by two other mechanisms: the non-dominated sorting, and the crowding sorting.

\begin{definition}
\label{def:non-dominated-sorting}
A \textbf{non-dominated sorting} of a population $P$ corresponds to sorting each individual into a non-domination class. The first domination class $F_1$ is composed of the non-dominated individuals of $P$, the second non-domination class $F_2$ is composed of the non-dominated individuals of $P$ without the individuals in $F_1$, and so on, until no individual is left in $P$.
\end{definition}

\begin{figure}
\center
\begin{tabular}{cc}
    \includegraphics[width=0.325\linewidth]{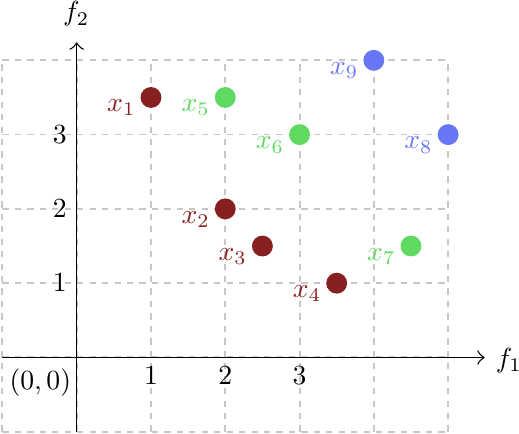} & \includegraphics[width=0.625\linewidth]{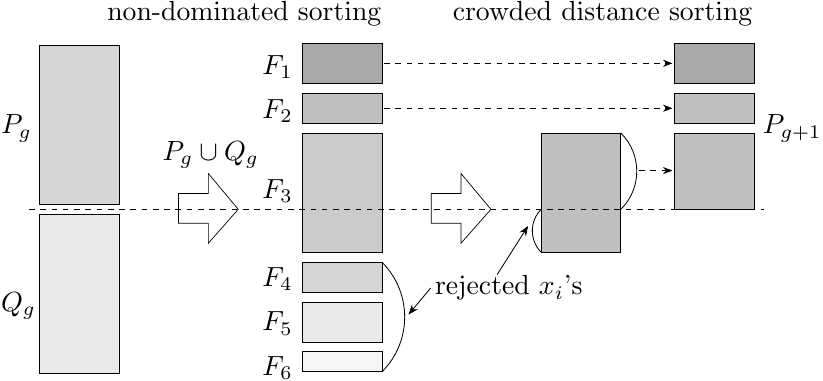} \tabularnewline
    (a) & (b) \tabularnewline
\end{tabular}
\caption{(a) Example of a non-dominated sorting. Objectives $f_1$ and $f_2$ need to be minimized. Solutions in red, green, and blue, represent the first, second, and third non-domination classes, respectively. (b) Illustration of selection of a new population in NSGA-II. $R_g$ is the union of $P_g$ (population in generation $g$) and $Q_g$ (varied individuals in generation $g$). In this example, non-domination classes $F_1$ and $F_2$ go into new population of generation $g+1$, $P_{g+1}$. Adding the complete non-domination class $F_3$ would exceed the population size, thus individuals from $F_3$ with higher crowding distance are selected. }
\label{fig:nsga-ii}
\end{figure}

Figure \ref{fig:nsga-ii}-(a) illustrates the concepts of Pareto front and non-domination classes. 

\begin{definition}
\label{def:crowded-sorting}
The \textbf{crowded sorting} of a non-domination class is the sorting of individuals in descending order of the crowding distance. The crowding distance of an individual is the volume in the objective space around it that is not covered by any other solution, calculated as the perimeter of the cuboid that has the nearest neighbors in the objective space as vertices.
\end{definition}

\begin{definition}
\label{def:crowded-comparison}
Given two individuals $x_i$ and $x_j$, with non-domination classes $x_i^{rank}$ and $x_j^{rank}$, and crowding distances $x_i^{dist}$ and $x_j^{dist}$, respectively, the \textbf{crowded comparison operator} $\prec_c$ defines the partial order $x_i \prec_c x_j$ if and only if ($x_i^{rank} < x_j^{rank}$) or ($x_i^{rank} = x_j^{rank}$ and $x_i^{dist} > x_j^{dist}$).
\end{definition}

By considering both the non-dominated sorting and the crowded sorting in the crowded comparison operator, NSGA-II keeps balances elitism and preservation of diversity of individuals inside a Pareto front. More detailed explanations of these metrics can be found in \citet{deb2002nsgaii}. With these definitions at hand, the NSGA-II algorithms differs from a standard genetic algorithm in: 1) The crowded comparison operator is used when comparing individuals for selection in tournament, and 2) Given a population $P$, at each generation a set $Q$ of offspring is generated via selection and application of genetic operators. The union of $P$ and $Q$ is then sorted in non-domination classes and each class is internally sorted according to the crowding distances. The non-domination classes are then added to the next population until the population size is reached. If including the last non-dominated class before reaching the population size exceeds the population size, then individuals are chosen according to their crowding distance, in decreasing order. The process is illustrated in figure \ref{fig:nsga-ii}-(b). Algorithm \ref{alg:nsga-ii} summarizes the optimization procedure of NSGA-II.

\begin{algorithm2e}
    \SetAlgoLined
    \DontPrintSemicolon
    \KwData{population size $N$ and maximum number of generations $max\_gens$}
    \KwResult{population $P_g$ of size $N$}
     sample and evaluate initial population $P_0$ of size $N$\;
     $g \gets 0$    \;
    \While{$g < max\_gens$ and quality criterion not met}{
    offspring set $Q_g \gets \emptyset$\;
        \While{$|Q_g| < N$}{
            select individuals $x_1$ and $x_2$ via tournament and crowded comparison operator $\prec_c$\;
            apply crossover and mutation\;
            $Q_g \gets Q_g \cup \{x_1, x_2\}$ \tcp*{add offspring $x_1$ and $x_2$ to $Q_g$}    
        }
        $R_g \gets P_g \cup Q_g$\;
        order $R_g$ according to non-dominated and crowded ordering to obtain $F_i$'s\;
        $P_{g+1} \gets \emptyset$\;
        $i \gets 0$\;
        \While{$|P_{g+1}| < N$}{
            \eIf{$|F_i| + |P_{g+1}| \leq N$}{
            add non-domination class $F_i$ to $P_{g+1}$
            }{
            add first $N - |P_{g+1}|$ individuals from non-domination class $F_i$ to $P_{g+1}$
            }
            $i \gets i + 1$
        }
        $g \gets g + 1$
    }
    \Return{$P_g$}
\caption{Pseudocode for NSGA-II.}
\label{alg:nsga-ii}
\end{algorithm2e}

% \section{Methodology}

\section{Definition of the Flexibility Benchmark}
\label{sec:methodology}
\label{sec:definition-task}

Following section~\ref{sec:system-flexibility}, we now describe the details of the flexibility benchmark for the multi-objective optimization of process parameters based on the extended Oxley model presented in section \ref{sec:oxley-model}. The task context consists of four related multi-objective metal cutting process optimization tasks $\mathcal T = \{T_1, T_2, T_3, T_4\}$, where each task is associated to a different type of metal to be cut. We consider four different materials: \textit{steel}, \textit{tungsten alloy}, \textit{steel dummy}, and \textit{inconel-718}. In the extended Oxley model, a material is characterized by the \textit{material parameters} described in table \ref{tab:material-params}. The specific material parameters for each of the four considered metal types is shown in table \ref{tab:tasks-materials}.

In each task, we assume that a certain \textit{total length} (in $m$) and a certain \textit{total depth} (in $mm$) of material has to removed. We keep the total length and total depth fixed across all tasks. The \textit{process parameters} that span the solution space $X$ are cutting speed, cutting angle, and cutting depth. Table \ref{tab:process-params} provides a description of these parameters as well as a suggested range, based on preliminary experimentation with the extended Oxley model. Given a specification of process parameters, the model outputs the values shown in table \ref{tab:outputs}. We use a combination of the process parameters and the output values to define a \textit{feasibility criterion} and a \textit{performance measure} for candidate solutions. A solution is feasible if the cutting speed is below $50 m/s$ and the output forces $F_c$ and $F_t$ are both below $500N$. The performance of a solution is measured by four objectives that should be minimized: \textit{production time}, \textit{tool wear}, and the absolute output forces $F_c$ and $F_t$. The production time, measured in seconds, is the time needed for removing the material, and is defined as:
\begin{align}
\label{eq:production-time}
    production\_time = (total\_length / cutting\_speed) * n\_layers.
\end{align}

The tool wear measures how much the tool is affected by the operation, and is defined as:
\begin{align}
\label{eq:tool-wear}
    tool\_wear = (cutting\_speed * e^{|F_c|} + 0.1 * cutting\_speed * e^{|F_t|}) * n\_layers.
\end{align}

As minimizing these four objectives can be contradictory at times, the optimization consists of finding a Pareto front of solutions, from which a domain expert can choose the most suitable solution at a given time. Based on the ranges for the input process parameters (table \ref{tab:process-params}), the maximum threshold for the output forces $F_c$ and $F_t$, and equations \ref{eq:production-time} and \ref{eq:tool-wear}, we present in table \ref{tab:objectives} the ranges that the four objectives can achieve.

\begin{table}[b]
\center \small
\caption{Description of material parameters that define a material and a task. From \cite{pantale-2022} and their provided repository.}
\begin{tabular}{c|c|c}
    \hline
    \textbf{Parameter} & \textbf{Description} & \textbf{Unit} \tabularnewline
    \hline
    $T_0$                  & Initial temperature.                                     & Kelvin     \tabularnewline
    $T_w$                  & Ambient temperature.                                     & Kelvin     \tabularnewline
    $\rho$                 & Density.                                                 & $km/m^3$   \tabularnewline
    $\eta$                 & Temperature averaging factor for shear plane.            & -          \tabularnewline
    $\psi$                 & Temperature averaging factor for tool-chip interface.    & -          \tabularnewline
    $jc\_A$                & $A$ coefficient in the Johnson-Cook law.                 & Pa         \tabularnewline
    $jc\_B$                & $B$ coefficient in the Johnson-Cook law.                 & Pa         \tabularnewline
    $jc\_n$                & $n$ coefficient in the Johnson-Cook law.                 & -          \tabularnewline
    $jc\_C$                & $C$ coefficient in the Johnson-Cook law.                 & -          \tabularnewline
    $jc\_m$                & $m$ coefficient in the Johnson-Cook law.                 & -          \tabularnewline
    $T_m$                  & Melting temperature.                                     & Kelvin     \tabularnewline
    $jc\_\dot{\epsilon}_0$ & $\dot{\epsilon}_0$ coefficient for the Johnson-Cook law. & $1/s$      \tabularnewline
    \hline
\end{tabular}
\label{tab:material-params}
\end{table}

\begin{table}
\center \small
\caption{Material parameter specification for the four materials considered. Definition of material parameters in table \ref{tab:material-params}. Values for steel and tungsten alloy taken from \cite{rashed-2016}. Values for steel dummy and inconel-718 defined by the authors based on previous internal projects. Values for $\eta$ and $\psi$ kept constant and are taken from the implementation of the extended Oxley model provided by \citet{pantale-2022}.}
\begin{tabular}{c|cccc}
    \hline
     & \multicolumn{4}{c}{\textbf{Material}} \tabularnewline
    \textbf{Parameter} & \textbf{Steel} & \textbf{Tungsten Alloy} & \textbf{Steel Dummy} & \textbf{Inconel-718} \tabularnewline
    \hline
    $T_0$                  & 273.15    & 273.15   & 273.15   & 273.15   \tabularnewline
    $T_w$                  & 300       & 300      & 300      & 300      \tabularnewline
    $\rho$                 & 7,860     & 17,600   & 7,860    & 8,242    \tabularnewline
    $\eta$                 & 0.9       & 0.9      & 0.9      & 0.9      \tabularnewline
    $\psi$                 & 0.9       & 0.9      & 0.9      & 0.9      \tabularnewline
    $jc\_A$                & 7.92e+08  & 1.51e+09 & 5.82e+08 & 9.28e+08 \tabularnewline
    $jc\_B$                & 5.10e+08  & 1.77e+08 & 4.65e+08 & 9.79e+08 \tabularnewline
    $jc\_n$                & 0.26      & 0.12     & 0.325    & 0.245847 \tabularnewline
    $jc\_C$                & 0.014     & 0.016    & 0.008    & 0.0056   \tabularnewline
    $jc\_m$                & 1.03      & 1        & 1.3      & 1.80073  \tabularnewline
    $T_m$                  & 1,790     & 1,723    & 1,790    & 1,623.15 \tabularnewline
    $jc\_\dot{\epsilon}_0$ & 1         & 1        & 1        & 0.001    \tabularnewline
    \hline
\end{tabular}
\label{tab:tasks-materials}
\end{table}

\begin{table}
\center \small
\caption{Description of process parameters that serve as input for simulation. Cutting width was kept fixed, as it only changed the scale of the outputs. Taken from \cite{pantale-2022} and their provided repository. Suggested ranges calculated from preliminary experiments.}
\begin{tabular}{c|c|c|c}
    \hline
    \textbf{Parameter} & \textbf{Description} & \textbf{Unit} & \textbf{Suggested Range} \tabularnewline
    \hline
    $cutting\_speed$ & Tool cutting speed.       & $m/sec$ & 0.1 to 5.0       \tabularnewline
    $cutting\_angle$ & Tool rake angle.          & radians & -0.5 to 1.0      \tabularnewline
    $cutting\_width$ & Width of cut.             & $mm$    & Fixed = 1.6e-4   \tabularnewline
    $cutting\_depth$ & Depth of cut in one step. & $mm$    & 1.0e-6 to 1.0e-3 \tabularnewline
    \hline
\end{tabular}
\label{tab:process-params}
\end{table}

\begin{table}
\center \small
\caption{Description of simulation outputs. Taken from \cite{pantale-2022} and their provided repository. The number of layers needed to remove all material is calculate by $total\_depth/cutting\_depth$, meaning we would need $n\_layers$ steps. As the outputs are the same for each step, we take the ones from the last step to calculate the objectives.}
\begin{tabular}{c|c|c}
    \hline
    \textbf{Observation} & \textbf{Description} & \textbf{Unit} \tabularnewline
    \hline
    $shear\_angle$ & Angle at which chip separates from material during cutting. & radians \tabularnewline
    $F_c$          & Chip formation force in cutting direction.                  & $N$     \tabularnewline
    $F_t$          & Chip formation force in thrust direction.                   & $N$     \tabularnewline
    $t_c$          & Chip thickness.                                             & $mm$    \tabularnewline
    $n\_layers$    & Layers needed to remove all material.                       & -       \tabularnewline
    \hline
\end{tabular}
\label{tab:outputs}
\end{table}

\begin{table}[htb]
\center \small
\caption{Description of the objectives of a task.}
\begin{tabular}{c|c|c|c}
    \hline
    \textbf{Objective} & \textbf{Description} & \textbf{Unit} & \textbf{Achieved Range} \tabularnewline
    \hline
    $production\_time$ & Time needed for removing the material.     & $s$ & 200 to 10e6     \tabularnewline
    $tool\_wear$       & Damage to the tool.                        & -   & 110 to 7.72e223 \tabularnewline
    $F_c$              & Chip formation force in cutting direction. & $N$ & 0 to 500        \tabularnewline
    $F_t$              & Chip formation force in thrust direction.  & $N$ & 0 to 500        \tabularnewline
    \hline
\end{tabular}
\label{tab:objectives}
\end{table}

\section{NSGA-II as a Baseline for Evolutionary Adaption of Solutions}
\label{sec:nsga-ii-facilitated-variation}

As a baseline method for finding solutions for optimizing the objectives defined in section \ref{sec:definition-task}, we have chosen NSGA-II, which is an established algorithm for finding a Pareto front for multi-objective optimization problems. We represent solutions as a real-valued vector, where each position refers to one process parameter being evolved (cutting speed, cutting angle, and cutting depth, as in table \ref{tab:process-params}). In the context of adaption of solutions from source to target tasks, we performed the following modifications to standard NSGA-II:

\begin{enumerate}
    \item We run the algorithm for a given task for a fixed number of generations and save the best Pareto front found. For assessing the quality of a Pareto front, we measure its hypervolume, although other measures can also be used for that \cite{riqueline2015performance, wang2016practical, AUDET2021397}.

    \item For adaption, we take the stored Pareto front from a source task as initial population for optimizing for the target task. If the loaded Pareto front is lower than the population size, we complete the initial population with randomly generated individuals. However, the stored Pareto front was always as large as the population size in our experiments. In principle, adaption stops when a Pareto front is found with the same hypervolume as found from scratch for the target task.
\end{enumerate}

In a preliminary analysis, where we sampled 10,000 random solutions for each material (details for material parameters and process parameter ranges from which to sample in sections \ref{sec:definition-task} and  \ref{sec:setup-parameters}, respectively), we observed that the approximated Pareto front of different materials lie on the same region of the search space, although they differ in shape and exactly where. Based on this, we hypothesize that a Pareto front more in between two of such approximated Pareto fronts would be a better starting point for adaptation to a target material, although it might not be the best solution for a particular material. Here, we took inspiration from the works by \citet{parter-2008} and \citet{kashtan2007varying}, where concepts of facilitated variation are applied to a genetic algorithm that evolves circuits for goals that vary over time and as a result produces solutions that can be much more easily adapted to a target circuit that uses the same modules as the ones used for training. Further, we also extend the genotype used for accommodating two or more values for each process parameter, from which only one is active at a time. 
Although we take inspiration from the work by \citet{parter-2008}, these two proposed variants for adaption are also related to existing algorithms from the field of dynamic (multi-objective) evolutionary optimization \cite{branke2012evolutionary, yang2015dynamic, azzouz2017dynamic}.
We present these two extensions to NSGA-II in the next two sections.

\subsection{Varying Goals}
\label{sec:varying-goals}

The varying goals evolution strategy consists of optimizing in one run for two or more goals at the same time. The goal is varied according to a parameter that we here call \textit{epoch length E}; the goal is changed each $E$ generations. Thus, a same population is optimized for different goals that change each $E$ generations, which defines an epoch. Given $n$ goals $g_1, g_2, ..., g_n$, and an epoch length $E$, the goal index in generation $i$, when generation begins with 1, is calculated as:
\begin{align}
\label{eq:goal-idx}
    idx_{goal} = \lfloor(i-1)/E\rfloor \bmod n.
\end{align}

Optimizing for different goals raises the question of which Pareto front to store for further adaption. We store the best hypervolume found for each goal and update the \textit{best Pareto front so far} each time this best value for the \textit{current goal} improves. That is, the stored Pareto front is not necessarily the one that achieved the best global quality, but the one that last improved on the best value of a given current goal. We chose to do so because the hypervolume achieved for each material is different (see section \ref{sec:results-hypervolumes}). If we stored the Pareto front with the best global hypervolume, we could be favouring one goal over the others. By storing the one that last improves the best value of the current goal, we also ensure that the stored Pareto front has gone through more iterations of optimization.

\citet{parter-2008} propose the varying goals strategy in the biology context in order to study the mechanisms of \textit{facilitated variation}, specifically the elements of \textit{modularity} and \textit{weak regulatory linkage} (see section \ref{sec:system-flexibility}). They show that, when optimizing under different but modular goals, the solutions can be quickly adapted to one or the other goal, or to other goals composed of different combinations of the same module, by mutations that change the connections between learned modules. For the problem we consider, the genotype is a vector of three real-valued numbers that represent process parameters, so it is difficult to imagine modules in there, although one could still argue that solutions to different goals could share building blocks that can be swapped through crossover. In order to go beyond just making solutions stay in-between the optimal region for different materials, we also propose a way to accommodate different possibilities in one solution, using a representation model we call \textit{active-inactive genotype}, which we discuss in the next section.

The varying goals strategy relates to the actual task of dynamic optimization, although in the context of having better starting points for manufacturing optimization, where the topic of reducing the number of necessary evaluations is still open and recent approaches deal more with surrogate models for the simulations \cite{emmerich2018tutorial}. 

\subsection{Active-Inactive Genotype}
\label{sec:active-inactive genotype}

What we call an active-inactive genotype refers to a genotype with both active and inactive positions, where only active positions appear in the phenotype, and positions can be both activated or deactivated via mutations. In the context of evolutionary computation, the concept of inactive genes or nodes appears in specific representations for genetic programming \cite{TurnerNeutralDrifCGP, Sotto2021GraphRI}. In these representations, the interest for such mechanism lies mostly in neutral search, where solutions can escape local optima via neutral mutations, while some works also examine the hypothesis of evolved information that can be deactivated and further reactivated influencing search \cite{sottoNonEffectiveLGP}. Although this does not seem to always be the case when evolving for a fixed goal, we consider the hypothesis that, under varying goals, information evolved for one goal can be deactivated when the goals change but can be further activated for different goals or for adaption. That is, we propose studying the interplay between active and inactive positions as a way of the genotype being able to store information about different goals and thus using this for improving adaption to a target goal.

\begin{figure}
\center
\begin{tabular}{c}
    \includegraphics[scale=0.5]{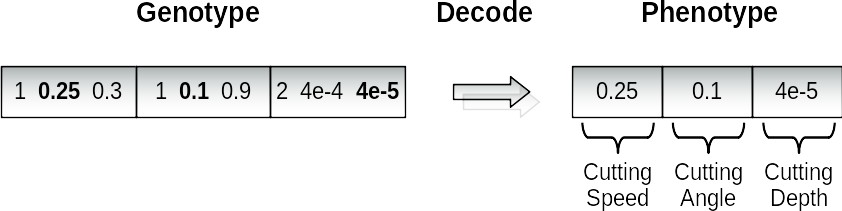} \tabularnewline
    \tabularnewline
    (a) \tabularnewline
    \tabularnewline
    \includegraphics[scale=0.5]{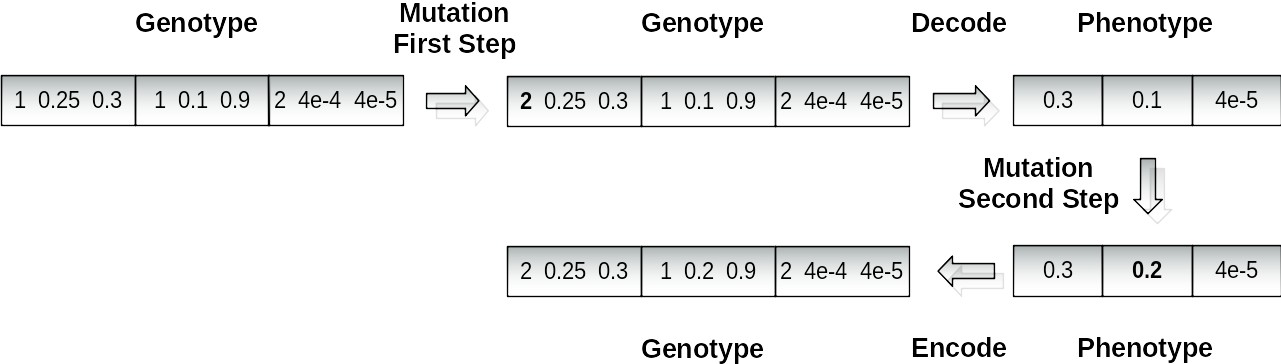} \tabularnewline
    \tabularnewline
    (b) \tabularnewline
\end{tabular}
\caption{(a) Example solution using the active-inactive genotype, with gene length $l=2$. The first position of a gene indicates which of the next $l$ positions is active. Active positions are in boldface. The decoded phenotype contains only the values in the active position for each gene. (b) Example of the application of the modified two-step mutation operator. First, indexes of active positions can be changed (in boldface after first step). Then, the individual is decoded, standard mutation is applied (boldface after second step), and the result is again encoded.}
\label{fig:example-active-inactive}
\end{figure}

In our proposed representation model, given a \textit{gene length l}, each gene, associated to a process parameter, is composed of $l+1$ positions. The first position tells which of the next $l$ positions is active; the next $l$ positions each encode a possible value for the given process parameter. A phenotype is derived by taking the active position of each gene, resulting in the real-valued vector given as input to the simulation for evaluation. Figure \ref{fig:example-active-inactive}-(a) shows an example solution and the decoded phenotype. Algorithm \ref{alg:decode} shows a pseudocode for decoding a solution.

We have also adapted the genetic operators to work with the proposed genotype. Crossover acts on the decoded phenotype as a regular crossover, then the result is encoded into the genotype. Mutation works with two steps. First, the first position of each gene can mutated to activate or deactivate  with a probability of $1/n_{process}$, where $n_{process}$ is the number of process parameters, so that one position is activated/deactivated on average. Next, the decoded phenotype is mutated as in a regular mutation, and the result is encoded in the genotype. A pseudocode for encoding a phenotype in a genotype is shown in algorithm \ref{alg:encode}, and algorithm \ref{alg:two-step-mutation} shows a pseudocode of the two-step mutation operator. Although shown separately in the algorithms, the decoded phenotype is always stored together with the encoded genotype, to avoid having to decode it again for evaluation. Figure \ref{fig:example-active-inactive}-(b) shows an example of the application of the modified two-step mutation operator.

\begin{algorithm2e}
    \SetAlgoLined
    \DontPrintSemicolon
    \KwData{$ind$ is the genotype to be decoded, $l$ is the gene length}
    \KwResult{decoded phenotype \textbf{decode(ind, l)}}
    decoded $\gets$ [], 
     $i \gets 0$\;
    \While{ i < length(\textit{ind})}{
        decoded.append(\textit{ind}[i+\textit{ind}[i]])\;
         $i \gets i + l + 1$
    }
    \Return{decoded}
\caption{Pseudocode for obtaining a decoded phenotype from a genotype with active and inactive positions.}
\label{alg:decode}
\end{algorithm2e}

\begin{algorithm2e}
    \SetAlgoLined
    \DontPrintSemicolon
    \KwData{\textit{ind} is the original encoded genotype, \textit{decoded} is a modified decoded phenotype, and $l$ is the gene length}
    \KwResult{encoded phenotype \textbf{encode(ind, decoded, l)}}
    \For{i from 0 to length(\textit{decoded})}{
        idx $\gets$ i * ($l$ + 1)\;
        \textit{ind}[idx+\textit{ind}[idx]] $\gets$ \textit{decoded}[i]\;
    }
    \Return{ind}
\caption{Pseudocode for encoding a phenotype given a genotype with active and inactive positions.}
\label{alg:encode}
\end{algorithm2e}

\begin{algorithm2e}
    \SetAlgoLined
    \DontPrintSemicolon
    \KwData{\textit{ind} is the encoded individual to be mutated, $l$ is the gene length, and $n_{process}$ is the number of process parameters}
    \KwResult{\textbf{mutate\_two\_steps(ind, l,} $\mathbf{n_{process}}$\textbf{)}}
    $i \gets 0$\;
    \While{ i < length(\textit{ind})}{
    \Case{with probability $1/n_{process}$}{
         \textit{ind}[i] = different random integer between 1 and l}
         $i \gets i + l + 1$
    }
    \textit{decoded} $\gets$ decode(\textit{ind}, l)\;
    mutate \textit{decoded}\;
    \textit{ind} = encode(\textit{ind}, \textit{decoded}, l)\;
    \Return{ind}
\caption{Pseudocode for the two-step mutation on the genotype with active and inactive positions.}
\label{alg:two-step-mutation}
\end{algorithm2e}

The active-inactive genotype relates to the concept of memory, which, in the dynamic evolutionary optimization literature, stores and reuses useful information from previous goals, and works well in cyclic environments \cite{yang2015dynamic, azzouz2017dynamic}. More specifically, the approach is related to the strategy of implicit memory, that uses redundant representations to store information, like in, for example, the diploid genetic algorithm from \citet{uyar2005diploid}. In the context of NSGA-II or memory for multi-objective dynamic problems, \citet{deb2007dynamic} introduce a dynamic NSGA-II using a diversity promotion strategy, \citet{goh2009coevolutionary} propose an algorithm with memory using coevolution of subpopulations, and \citet{wang2009memory} propose an NSGA-II with memory. However, the way we build an implicit memory via the active-inactive genotype differs from these previous works; moreover, we apply the strategy to train a solution that will be adapted to solve a previously unseen problem in the context of static manufacturing optimization, as opposed to a cyclic environment.

\subsection{Parameters and Implementation Details}
\label{sec:setup-parameters}

The base implementation used for the extended Oxley model was the one provided by \citet{pantale-2022}.\footnote{\url{https://github.com/pantale/OxleyPython}} For the implementation of the NSGA-II algorithm, we have used the DEAP library in Python \cite{DEAP_JMLR2012}. For tournament and selection of next population, we use the provided NSGA-II selection method\footnote{\url{https://deap.readthedocs.io/en/master/api/tools.html\#deap.tools.selNSGA2}}. As genetic operators, we use the simulated binary bounded crossover\footnote{\url{https://deap.readthedocs.io/en/master/api/tools.html\#deap.tools.cxSimulatedBinaryBounded}} and the polynomial bounded mutation\footnote{\url{https://deap.readthedocs.io/en/master/api/tools.html\#deap.tools.mutPolynomialBounded}}. The simulated binary bounded crossover swaps genes between two parent chromosomes and may apply a perturbation to some genes. The polynomial bounded mutation applies a perturbation to each gene with probability $1/n_{process}$, where $n_{process}$ is the number of process parameters being optimized and thus the chromosome length. Both operators respect lower and upper bounds for each gene, which are the lower and upper bound for the process parameters provided in table \ref{tab:process-params}. The extended algorithms with varying goals and active-inactive genotype were implemented on top of the baseline algorithm according to the explanations and pseudocodes from section \ref{sec:nsga-ii-facilitated-variation}.

Table \ref{tab:nsga-params} shows the parameter specifications for NSGA-II and the two proposed variants. In our preliminary analysis, where we sampled 10,000 random solutions to estimate the Pareto front for each material, Pareto fronts had approximately 100 to 300 solutions. We thus set the population size to 100 and the number of generations to 50. We use a standard tournament size of 2 based on preliminary runs. The $\eta_{cross}$ and $\eta_{mut}$ parameters for crossover and mutation, respectively, define the magnitutde of the perturbation to the chromosome values - high values produce offspring more similar to its parents, whereas low values produces more different offspring. We performed 50 runs using different value combinations (20, 40, 80, 120, 140, 180) for optimizing for steel, inconel-718, and from steel to inconel-718. We observed that low values produce Pareto fronts with higher hypervolumes, and thus use the suggested values from the DEAP documentation. We set the epoch length $E$ to not be too large or too small, and the gene length $l$ to not be too large. 
% We perform a further study on these parameters in section \ref{sec:results-epoch-gene-lengths}, and on the population size and maximum generations in section \ref{sec:results-lower-pops}.

\begin{table}
\center 
\caption{Parameters used for the NSGA-II algorithm. The epoch length $E$ applies only to the variants with varying goals and active-inactive genotype. The gene length $l$ applies only to the variant with active-inactive genotype.}
\begin{tabular}{c|c}
    \hline
    \textbf{Parameter} & \textbf{Value} \tabularnewline
    \hline
    Population Size     & 100 \tabularnewline
    Maximum Generations & 50  \tabularnewline
    Tournament Size     & 2   \tabularnewline
    $\eta_{cross}$      & 30  \tabularnewline
    $\eta_{mut}$        & 20  \tabularnewline
    Epoch Length $E$    & 5   \tabularnewline
    Gene Size $l$       & 2   \tabularnewline
    \hline
\end{tabular}
\label{tab:nsga-params}
\end{table}

\section{Results and Discussion}
\label{sec:results}

Using the benchmark described in section~\ref{sec:definition-task}, our analysis consists of comparing the best, average, and worst case adaption costs with the costs from scratch as obtained for each target material, and also comparing the adaption costs for the baseline NSGA-II, NSGA-II with varying goals, and NSGA-II with varying goals and active-inactive genotype (section \ref{sec:results-flexibility}).
To measure the quality of a Pareto front, we use the hypervolume, which is a measure of the volume in the objective space that is covered by the solutions in the Pareto front, given a reference point which is a vector of worst objective values \cite{riqueline2015performance, wang2016practical, AUDET2021397}. Thus, greater values stand for better Pareto fronts. To calculate the hypervolume, we take the objective values, with ranges described in table \ref{tab:objectives}, and first apply a natural logarithmic scale to production time and tool wear. We then normalize the values between 0 and 1. For normalization, we use as reference for lower and upper bounds for each objective the ranges provided in table \ref{tab:objectives}. We then take $(1,1,1,1)$ as the reference point representing the worst possible solution in the range from 0 to 1.

Besides comparing this measure of flexibility of different optimization algorithms, we also perform a study on the introduced parameters epoch length $E$ and gene length $l$, repeating a reduced flexibility experiment as the one described above for NSGA-II with varying goals and active-inactive genotype, in order to assess the influence of these parameters on optimization from scratch and adaption (section \ref{sec:results-epoch-gene-lengths}). Finally, we also repeat the flexibility experiment with different population sizes to study if the adaption cost can be reduced in situations where smaller populations are sufficient for finding a solution with the desired quality (section \ref{sec:results-lower-pops}).

% Section \ref{sec:results-flexibility} presents the results obtained for the baseline flexibility experiment, with a focus on comparing the flexibility of the three considered algorithms and how the proposed approaches reduce the adaption cost. Section \ref{sec:results-epoch-gene-lengths} presents the result of a study on the two introduced parameters epoch and gene length. In section \ref{sec:results-lower-pops}, we perform a study on the population size and repeat the flexibility experiment with lower population sizes in order to assess if the proposed methods provide an sustained advantage in different scenarios and if it is possible to further reduce the adaption cost.

\label{sec:setup-experiment}

\subsection{Flexibility Experiment}
\label{sec:results-flexibility}

\subsubsection{Obtained Hypervolumes}
\label{sec:results-hypervolumes}

For reference, table \ref{tab:reference-hypervolumes} shows the average maximum hypervolume obtained for each material over 100 runs when using the baseline NSGA-II with 100 individuals and 50 generations. As seen from the standard deviations, values are stable and vary only in the third or fourth decimal digit. The value achieved also depends on the material, ranging from 0.8813 for tungsten alloy to 0.9277 for steel dummy, which probably reflects properties of each task. These values found from scratch by the baseline NSGA-II are used as reference for adaption. That is, when adapting from steel to tungsten alloy, for example, the objective is to find a hypervolume for tungsten alloy similar to the one in table \ref{tab:reference-hypervolumes}.

Although some values may seem close to each other, as we take the logarithm of two objectives (production time and tool wear) and normalize all objectives between 0 and 1, a small difference in the hypervolume of a Pareto front may reflect a great difference in practice for the objective values. In this work, we chose to work with the hypervolume as an indicator of quality of a Pareto front and reference for adaption, but in more concrete problems, it should be possible for domain experts to define different goals.

\begin{table}
\center
\caption{Hypervolumes obtained from scratch for each material with the baseline NSGA-II (mean over 100 runs, standard deviation in brackets).}
\begin{tabular}{c|cccc}
    \hline
                         & \textbf{Steel} & \textbf{Tungsten Alloy} & \textbf{Steel Dummy} & \textbf{Inconel-718} \tabularnewline
    \hline
    \textbf{Hypervolume} & 0.9144 (7e-4)  & 0.8813 (1e-3)           & 0.9277 (7e-4)        & 0.8891 (1e-3)        \tabularnewline
    \hline
\end{tabular}
\label{tab:reference-hypervolumes}
\end{table}

\subsubsection{Hypervolumes Across Generations}
\label{sec:results-hypervolumes-gens}

\begin{figure}
\center
\includegraphics[width=0.95\columnwidth]{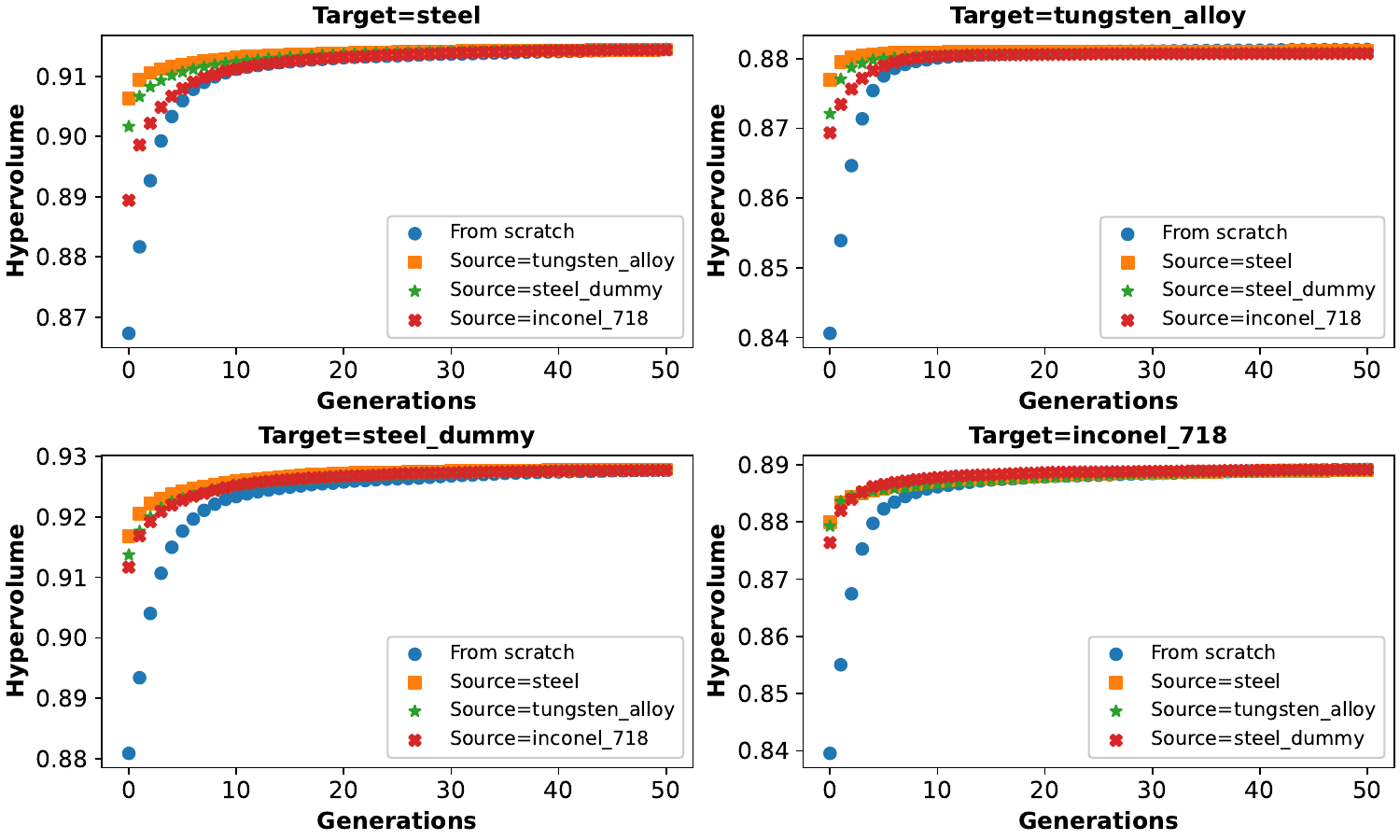}
\caption{Hypervolume of best Pareto front found so far across generations for search from scratch and adaption for each material (mean over 100 runs).}
\label{fig:hypervolumes-gens}
\end{figure}

In figure \ref{fig:hypervolumes-gens}, we show the hypervolume of the best Pareto front found so far across the generations averaged over 100 runs for each material, both for searching from scratch with the baseline NSGA-II, as for adapting from different source materials also with the baseline NSGA-II. For all cases, there is rapid improvement in the first 10 generations followed by phases of smaller improvements. Based on a visual inspection of the plots, when adapting from different source materials, the quality of the source Pareto front evaluated on the target task is already superior than a random initial population, and achieve higher values much quicker in comparison to search from scratch for more or less 10 generations, beyond which the hypervolumes become very similar both for search from scratch as well as for adaption. This enables us to state:

\begin{enumerate}
    \item Solutions from a source material work better than random solutions in a target material, and;
    
    \item Adapting solutions from a source material enables faster optimization towards a threshold quality.
\end{enumerate}

\begin{figure}
\center
\includegraphics[width=0.95\columnwidth]{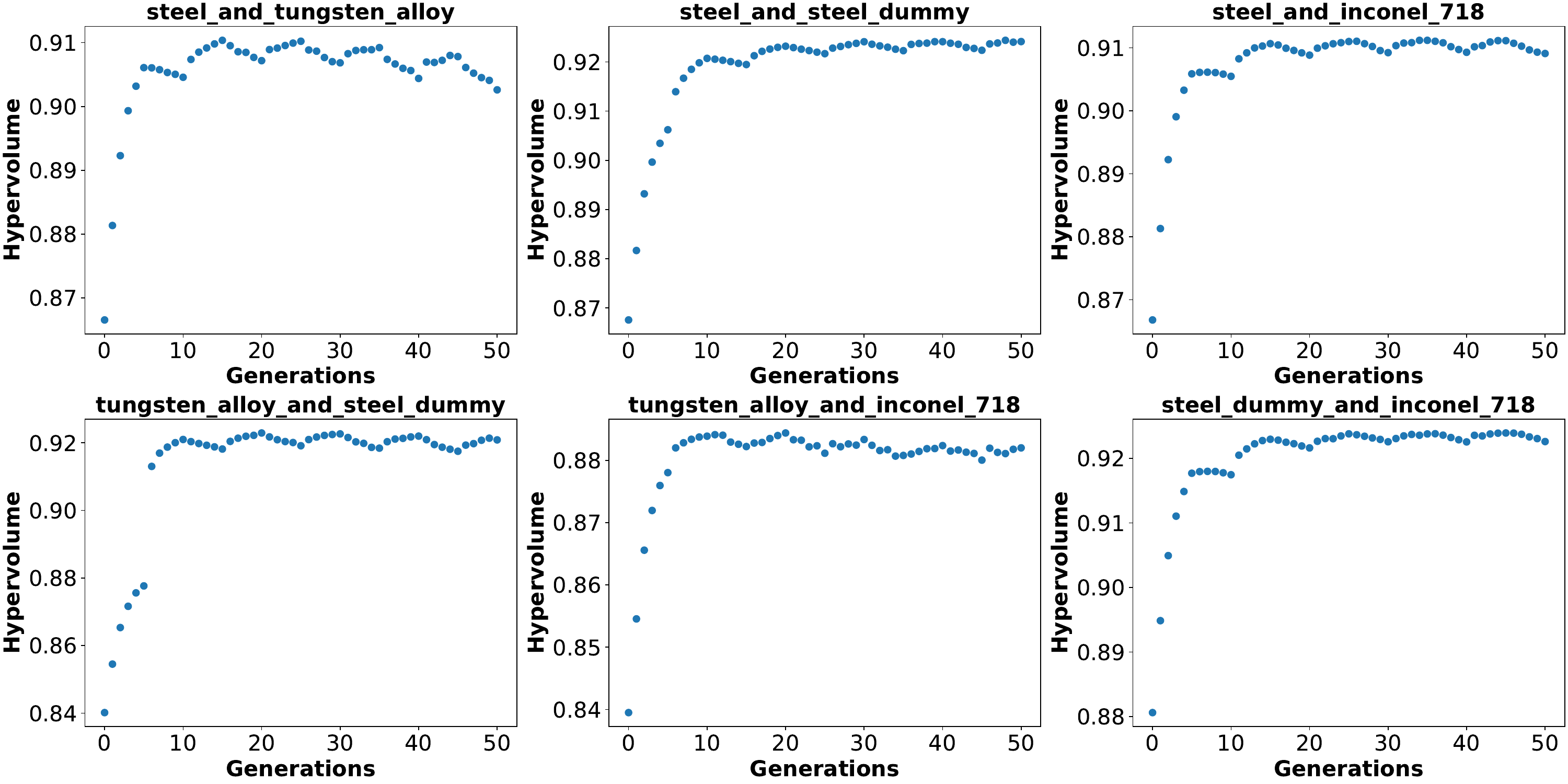}
\caption{Hypervolume of Pareto front found across generations for NSGA-II with varying goals for each material pair (mean over 100 runs). Here, $E=5$. For each epoch, the hypervolume shown correspond to the Pareto front evaluated on the objective \textit{of that epoch}, and not the best one found so far.}
\label{fig:hypervolumes-gens-varying-goals}
\end{figure}

We also show in figure \ref{fig:hypervolumes-gens-varying-goals} how the hypervolume develops across generations when using NSGA-II with varying goals, only for search from scratch, in order to visualize the effect of changing the goals each $E$ generations. Differently from figure \ref{fig:hypervolumes-gens}, here we show the hypervolume at each generation for the current goal, and not the best hypervolume found so far. As with the plots for the baseline NSGA-II, we observe a rapid increase in hypervolume in the first 10 generations. After that, although they remain more stable than in the beginning, we observe a different pattern each time the goal changes.\footnote{Plots for NSGA-II with varying goals and active-inactive genotype present the same pattern.}\textsuperscript{,}\footnote{Plots for baseline NSGA-II with the current hypervolume of each generation instead of the best so far presents frequent decreases in hypervolume only for tungsten alloy.} Interestingly, in almost in all cases the hypervolume for one goal always increases while it decreases for the other goal. This may be a effect of the solutions moving from one goal to another, but warrants further analysis. As already observed by \citet{parter-2008}, the quality of the solutions found under varying goals is not necessarily the best for a given goal, as what we aim at here is having an optimized population that lies in-between goals and is, thus, more adaptable, as we show next.

\subsubsection{Learning and Adaption Costs}
\label{sec:results-costs}

In our experiments, we found that having a hard threshold of only stopping adaption when the same reference hypervolume for the target material as shown in table \ref{tab:reference-hypervolumes} is found leads to adaption only succeeding in about 50\% of the cases. That is, at least in this problem, search from scratch probably benefits from the greater diversity in initial random populations to find solutions that are slightly better than what is possible when starting from solutions from a source material. Therefore, as stopping criterion for adaption we take 99\% of the reference hypervolumes found from scratch. That means that adaption stops when a Pareto front with 99\% of the reference hypervolume found from scratch for the target is found. As most applications that rely on costly simulations for evaluations cannot afford searching until the best possible solution is found and must adhere to a threshold defined by domain experts, we find the criterion of 99\% of the best possible hypervolume reasonable. Furthermore, when comparing adaption against search from scratch, we compare against the cost for finding 99\% of the reference hypervolume both for search from scratch as well as for adaption in each run, in order to highlight the differences in performance when aiming at the same threshold.

In our analysis in table \ref{tab:CE-all}, we compare the cost for finding 99\% of the reference hypervolume from scratch against the cost for adapting from different source materials. We show the Minimum Computational Effort (CE), which is a statistic that, given the number of evaluations needed for finding the solution in each run, estimates the minimum number of evaluations needed for finding the solution with a probability of 99\% \cite{Koza92}. As we are using a population size of 100 individuals, the minimum cost here is 100, when the loaded population already works.

From table \ref{tab:CE-all}, in general, already with the baseline NSGA-II the adaptive scheme reduces the cost for finding the threshold solution, with the exception of adaption from inconel-718 to steel, that has a cost of 1,000 in comparison to 900 for search from scratch. For NSGA-II with varying goals, we used pairs of goals to produce a source population, and the costs for adaption further decrease. Now there is one case (adaption from from steel dummy and inconel 718 to steel) where the cost is the same (900). Finally, NSGA-II with varying goals and active-inactive genotype obtains lower costs for adaption in comparison to search from scratch for all cases, including adaption from steel dummy and inconel-718 to steel, which now costs 600 evaluations.

\begin{table}
\center 
\caption{Minimum Computational Effort (CE) for search from scratch and for adaption, for baseline NSGA-II, NSGA-II with varying goals, and NSGA-II with varying goals and active-inactive genotype (CE calculated over 100 runs). Cells where the source is the same as the target are filled with "-".}
\begin{tabular}{c|c|cccc}
    \hline
    \multicolumn{6}{c}{\textbf{Baseline NSGA-II}} \tabularnewline
    \hline
      &  & \multicolumn{4}{c}{\textbf{Adaption From}} \tabularnewline
    \textbf{Target} & \textbf{From Scratch} & \textbf{Steel} & \textbf{Tungsten Alloy} & \textbf{Steel Dummy} & \textbf{Inconel-718} \tabularnewline
    \hline
    \textbf{Steel (S)}          & 900   & -   & 200 & 800 & 1,000 \tabularnewline
    \textbf{Tungsten Alloy (TA)} & 600   & 100 & -   & 200 & 600   \tabularnewline
    \textbf{Steel Dummy (SD)}    & 1,000 & 300 & 500 & -   & 700   \tabularnewline
    \textbf{inconel-718 (I)}    & 900   & 400 & 400 & 500 & -     \tabularnewline
    \hline
\end{tabular}
\vskip 0.1cm
\begin{tabular}{c|c|cccccc}
    \hline
    \multicolumn{8}{c}{\textbf{NSGA-II with varying goals}} \tabularnewline
    \hline
      &  & \multicolumn{4}{c}{\textbf{Adaption From}} \tabularnewline
      \textbf{Target} & \textbf{From Scratch}  & \textbf{S / TA} & \textbf{S / SD} & \textbf{S/ I} & \textbf{TA / SD} & \textbf{TA / I} & \textbf{SD / I} 
      \tabularnewline
    % \textbf{Target} & \textbf{From Scratch} & \textbf{Tungsten Alloy} & \textbf{Steel Dummy} & \textbf{inconel-718} & \textbf{Steel Dummy} & \textbf{inconel-718} & \textbf{inconel-718} \tabularnewline
    \hline
    \textbf{Steel (S)}          & 900   & -   & -   & -   & 600 & 500 & 900 \tabularnewline
    \textbf{Tungsten Alloy (TA)} & 600   & -   & 200 & 200 & -   & -   & 200 \tabularnewline
    \textbf{Steel Dummy (SD)}    & 1,000 & 300 & -   & 300 & -   & 400 & -   \tabularnewline
    \textbf{inconel-718 (I)}    & 900   & 400 & 500 & -   & 400 & -   & -   \tabularnewline
    \hline
\end{tabular}
% \resizebox{1\columnwidth}{!}{
% \begin{tabular}{c|c|cccccc}
%     \hline
%     \multicolumn{8}{c}{\textbf{NSGA-II with varying goals}} \tabularnewline
%     \hline
%       &  & \multicolumn{4}{c}{\textbf{Adaption From}} \tabularnewline
%       &  & \textbf{Steel and} & \textbf{Steel and} & \textbf{Steel and} & \textbf{Tungsten Alloy and} & \textbf{Tungsten Alloy and} & \textbf{Steel Dummy and} \tabularnewline
%     \textbf{Target} & \textbf{From Scratch} & \textbf{Tungsten Alloy} & \textbf{Steel Dummy} & \textbf{inconel-718} & \textbf{Steel Dummy} & \textbf{inconel-718} & \textbf{inconel-718} \tabularnewline
%     \hline
%     \textbf{Steel}          & 900   & -   & -   & -   & 600 & 500 & 900 \tabularnewline
%     \textbf{Tungsten Alloy} & 600   & -   & 200 & 200 & -   & -   & 200 \tabularnewline
%     \textbf{Steel Dummy}    & 1,000 & 300 & -   & 300 & -   & 400 & -   \tabularnewline
%     \textbf{inconel-718}    & 900   & 400 & 500 & -   & 400 & -   & -   \tabularnewline
%     \hline
% \end{tabular}}
% \vskip 0.1cm
\vskip 0.1cm
\begin{tabular}{c|c|cccccc}
    \hline
    \multicolumn{8}{c}{\textbf{NSGA-II with varying goals and active-inactive genotype}} \tabularnewline
      &  & \multicolumn{4}{c}{\textbf{Adaption From}} \tabularnewline
      \textbf{Target} & \textbf{From Scratch} & \textbf{S / TA} & \textbf{S / SD} & \textbf{S/ I} & \textbf{TA / SD} & \textbf{TA / I} & \textbf{SD / I} 
      \tabularnewline
    %   \hline
    %   &  & \multicolumn{4}{c}{\textbf{Adaption From}} \tabularnewline
    %   &  & \textbf{Steel and} & \textbf{Steel and} & \textbf{Steel and} & \textbf{Tungsten Alloy and} & \textbf{Tungsten Alloy and} & \textbf{Steel Dummy and} \tabularnewline
    % \textbf{Target} & \textbf{From Scratch} & \textbf{Tungsten Alloy} & \textbf{Steel Dummy} & \textbf{inconel-718} & \textbf{Steel Dummy} & \textbf{inconel-718} & \textbf{inconel-718} \tabularnewline
    \hline
    \textbf{Steel (S)}          & 900   & -   & -   & -   & 300 & 300 & 600 \tabularnewline
    \textbf{Tungsten Alloy (TA)} & 600   & -   & 200 & 100 & -   & -   & 200 \tabularnewline
    \textbf{Steel Dummy (SD)}    & 1,000 & 400 & -   & 400 & -   & 400 & -   \tabularnewline
    \textbf{inconel-718 (I)}    & 900   & 300 & 500 & -   & 300 & -   & -   \tabularnewline
    \hline
\end{tabular}
% \resizebox{1\columnwidth}{!}{
% \begin{tabular}{c|c|cccccc}
%     \hline
%     \multicolumn{8}{c}{\textbf{NSGA-II with varying goals and active-inactive genotype}} \tabularnewline
%     \hline
%       &  & \multicolumn{4}{c}{\textbf{Adaption From}} \tabularnewline
%       &  & \textbf{Steel and} & \textbf{Steel and} & \textbf{Steel and} & \textbf{Tungsten Alloy and} & \textbf{Tungsten Alloy and} & \textbf{Steel Dummy and} \tabularnewline
%     \textbf{Target} & \textbf{From Scratch} & \textbf{Tungsten Alloy} & \textbf{Steel Dummy} & \textbf{inconel-718} & \textbf{Steel Dummy} & \textbf{inconel-718} & \textbf{inconel-718} \tabularnewline
%     \hline
%     \textbf{Steel}          & 900   & -   & -   & -   & 300 & 300 & 600 \tabularnewline
%     \textbf{Tungsten Alloy} & 600   & -   & 200 & 100 & -   & -   & 200 \tabularnewline
%     \textbf{Steel Dummy}    & 1,000 & 400 & -   & 400 & -   & 400 & -   \tabularnewline
%     \textbf{inconel-718}    & 900   & 300 & 500 & -   & 300 & -   & -   \tabularnewline
%     \hline
% \end{tabular}}
\label{tab:CE-all}
\end{table}

Table \ref{tab:CE-all-summary} offers a more summarized view on these results, presenting the worst, average, and best, cases for search from scratch, adapting with the baseline NSGA-II, NSGA-II with varying goals, and NSGA-II with varying goals and active-inactive genotype. All costs decrease incrementally as we add features to the standard algorithm. These results confirm that optimization under varying goals generate solutions that are more adaptable, and that adding further structures to a genotype that allow it to store more information about different goals/options in a more complex way further enhances this effect. This happens because solutions for different materials, in this case, share some properties, like laying in a similar region of the search space.

We present the results of an analysis with non-parametric statistical tests as suggested by \citet{JMLR:v7:demsar06a}. We first applied the Friedman test for multiple algorithms on multiple data on the CE values from table \ref{tab:CE-all}. The Friedman test ranks the algorithms and calculates a \textit{p-value} which, if less than 0.05 (significance level of 95\%), means that there is a statistical difference between at least one pair of methods \cite{friedman1937ranks}. We obtain a \textit{p-value} of 1e-4. The post-hoc Nemenyi test then provides \textit{p-values} for each pair of methods \cite{nemenyi1963distribution}. The \textit{p-values} point to a difference between the baseline NSGA-II and the three variants for adaption, with incrementally lower values (0.036 when compared to baseline adaption, 0.004 to varying goals, and 0.001 to active-inactive genotype).

%As an interesting observation, NSGA-II with varying goals and active-inactive genotype presents adaption costs that range from 100 to 600. With a population size of 100 and considering that the first 100 evaluations correspond to loading the initial source population, this means that we need from 0 to 5 generations, that is, with a maximum of 5 steps of mutations, the algorithm is able to adapt source solutions to achieve a threshold hypervolume. This is in accordance to what is observed by \citet{parter-2008}: evolution of structures that can be activated or deactivated under varying goals leads to the manifestation of some characteristics of facilitated variation in living organisms, which possess a design that requires fewer mutations for adapting to changes in environments \cite{gerhart2007theory}.

\begin{table}
\center 
\caption{Worst, average, and best, cases for search from scratch and adaption, for baseline NSGA-II, NSGA-II with varying goals, and NSGA-II with varying goals and active-inactive genotype (calculated from values in table \ref{tab:CE-all}).}
\begin{tabular}{c|ccc}
    \hline
    \textbf{Algorithm} & \textbf{Worst Case} & \textbf{Average Case} & \textbf{Best Case} \tabularnewline
    \hline
    \textbf{From Scratch: Baseline}                    & 1,000 & 850 & 600 \tabularnewline
    \textbf{Adaption: Baseline}                        & 1,000 & 475 & 100 \tabularnewline
    \textbf{Adaption: Varying Goals}                   & 900   & 408 & 200 \tabularnewline
    \textbf{Adaption: Varying Goals + Active-Inactive} & 600   & 333 & 100 \tabularnewline
    \hline
\end{tabular}
\label{tab:CE-all-summary}
\end{table}

\subsection{Influence of Epoch and Gene Lengths}
\label{sec:results-epoch-gene-lengths}

As stated in section \ref{sec:setup-parameters}, we found it reasonable to use an epoch length $E=5$ that not so large in order to allow enough variation of goals and a gene length $l=2$ that is compatible with the fact that we considered only pairs or tasks as source for adaption. In order to have a better understanding of how these two introduced parameters influence both optimization from scratch as well as adaption, we performed a reduced flexibility experiment with different combinations of $E$ in $(5, 10, 15, 20, 25)$ and $l$ in $(2, 3, 4, 5)$. The experiments were performed for the pair of tasks steel and tungsten alloy and for the adaption to inconel-718, using again the hypervolumes in table \ref{tab:reference-hypervolumes} as reference for the threshold to be achieved (99\%).

\begin{figure}
\center
\includegraphics[width=0.95\columnwidth]{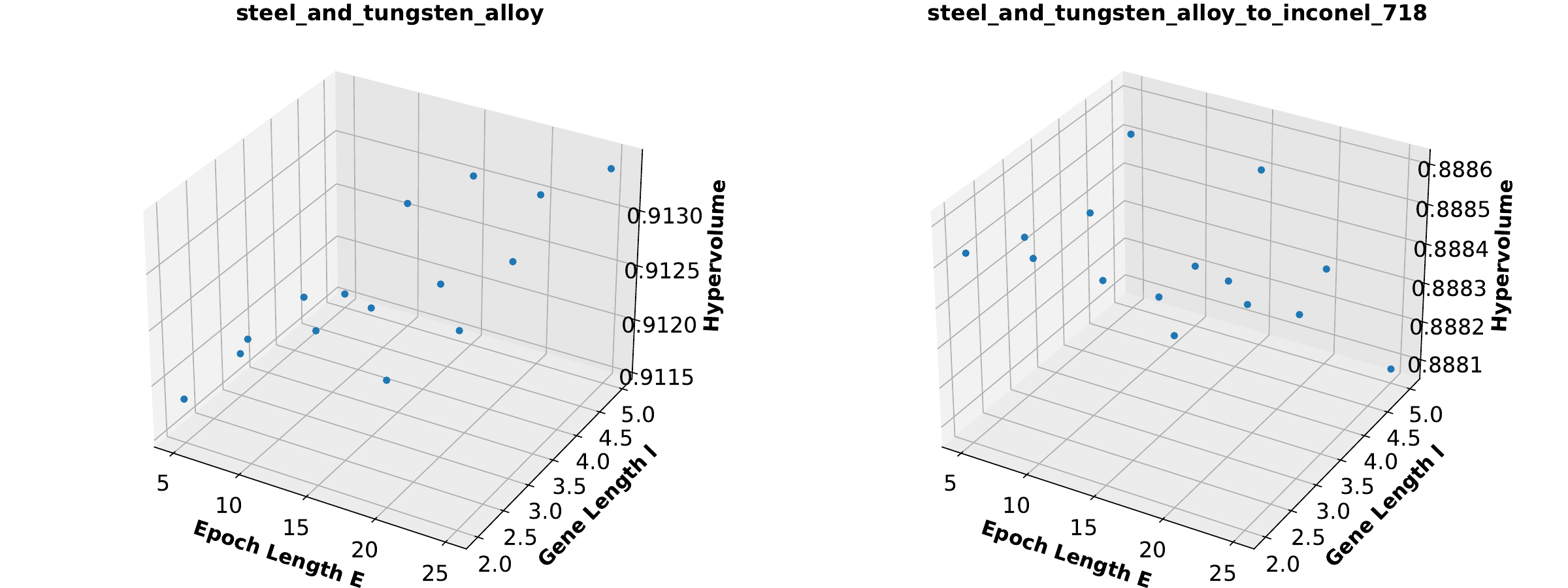}
\caption{Hypervolume of best Pareto front found for different combinations of epoch and gene lengths, for the pair steel and tungsten alloy, and for adaption to inconel-718 (values are a mean over 50 runs). For the pair of materials, the hypervolume shown is the best found in a run, regardless of goal.}
\label{fig:hypervolumes-epoch-gene-length}
\end{figure}

Figure \ref{fig:hypervolumes-epoch-gene-length} shows how the average best hypervolumes achieved over 50 runs vary in accordance to the parameters $E$ and $l$ for both considered tasks. We observe that the hypervolumes vary only slightly for search from scratch and even less for adaption, which makes sense, as in adaption we stop the search once the threshold hypervolume is achieved. Nonetheless, we can still discern two opposite behaviors. In search from scratch, optimization is better with longer epochs, which is expected as it gives more time for optimizing for a single goal. It is also interesting to notice that, with shorter epochs, a small gene size is preferred, whereas a larger gene size works better in combination with longer epochs. One possible explanation for that is that a larger gene size introduces more diversity that can be better explored in longer epochs for a fixed goal. Conversely, adaption works slightly better when search from scratch is worse (shorter epochs and smaller gene size), and slightly worse when search from scratch is better (longer epochs and larger gene size). By forcing more variation of goals, shorter epochs avoid over-optimization for a fixed goal. As for the gene size, we believe results could differ when optimizing for more than two goals at the same time, but the fact that larger gene sizes combined with longer epochs allow better optimization for one goal may also influence adaption negatively in the case under study.

\subsection{Experimenting with Lower Population Sizes}
\label{sec:results-lower-pops}

Although the proposed adaptive scheme and variants for NSGA-II were able to reduce the cost for finding a threshold Pareto front, as shown in section \ref{sec:results-costs}, the number of simulations reported are still high in the context of manufacturing optimization that requires the execution of costly simulations for evaluation of candidate solutions. Therefore, we explore in this section how the adaptive scheme and the proposed variants of NSGA-II behave when we use lower population sizes, both for search from scratch and adaption.

\begin{figure}
\center
\begin{tabular}{c}
    \includegraphics[width=0.95\columnwidth]{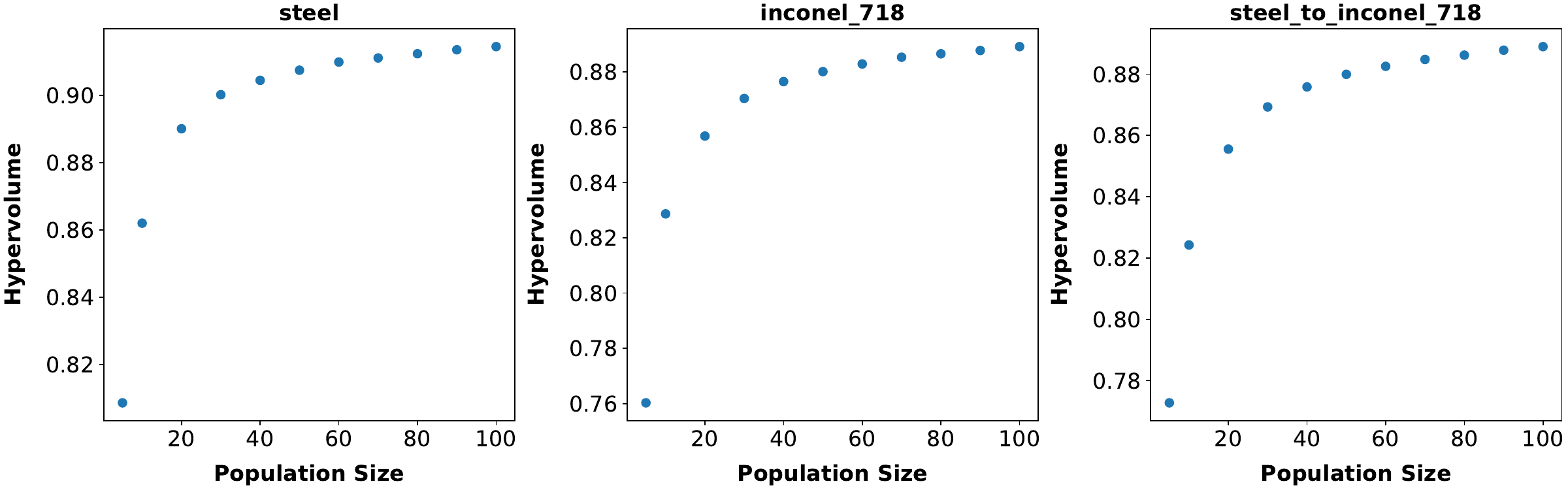} \tabularnewline
    \includegraphics[width=0.95\columnwidth]{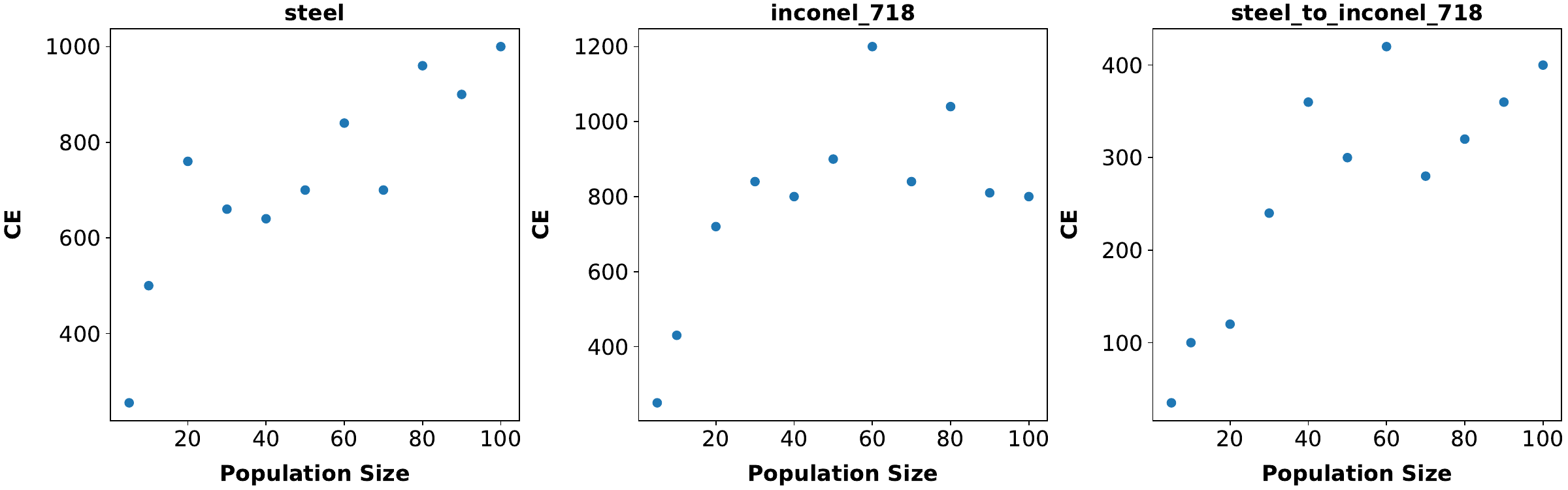} \tabularnewline
\end{tabular}
\caption{Hypervolume of Pareto front (top) and CE (bottom) for different population sizes, for steel, inconel-718, and adaption from steel to inconel-718 (values for hypervolume are a mean over 50 runs).}
\label{fig:hypervolumes-CE-pop-sizes}
\end{figure}

First, we ran a reduced flexibility experiment with population sizes 5 and 10 to 100 with a step of 10, kept the same number of generations as before (50), where we optimized from scratch for steel and inconel-718 and then adapted from steel to inconel-718, having as threshold 99\% of the hypervolume obtained from scratch for inconel-718, repeated 50 times. Figure \ref{fig:hypervolumes-CE-pop-sizes} shows the best hypervolumes and costs obtained. The CE shown both for adaption as well as for search from scratch was calculated considering the threshold of 99\% of the best hypervolume found in each run.

From figure \ref{fig:hypervolumes-CE-pop-sizes}, there is a decrease in best hypervolume obtained with every decrease in population size, with the value decreasing much more for population sizes 10 and 5. Accordingly, the cost for finding 99\% of this obtained hypervolume increases more or less linearly with the population size. There is an exception for inconel-718, where the cost remains more or less stable after population size 30. It can be that higher population sizes in some cases lead to better Pareto fronts more quickly, with less need for further generations, something we also observe in the results below. The cost for adaption also decreases with the population size, which makes sense as we are aiming at a lower target hypervolume. These results initially support the idea that, when a lower population size is enough for finding a desired solution, adaption would still help to reduce the computational cost.

Next, we repeat the same flexibility experiment as in section \ref{sec:results-flexibility} with all materials and algorithms and 100 runs, but with population sizes 50 and 20, first with 50 generations, and then with 100 generations for population size 50 and 250 generations for population size 20, in order to have the same number of evaluations (5,000) as with population size 100 and 50 generations from section \ref{sec:results-flexibility}. The motivation for having the same number of evaluations is to asses if, starting from a source that was optimized for more generations leads to better adaption as a trade-off for the longer training phase. The best hypervolumes obtained by the standard NSGA-II from scratch for each setup of population size and maximum generations are presented in table \ref{tab:reference-hypervolumes-pops-gens}, as an average over 100 runs. For all materials, one can observe that: 1) the obtained hypervolume is lower with lower population sizes, and 2) optimizing for more generations leads to a slight increase in hypervolume.

\begin{table}
\center 
\caption{Hypervolumes obtained from scratch for each material with the baseline NSGA-II using different population sizes and number of generations (mean over 100 runs, standard deviation in brackets).}
\begin{tabular}{cc|cccc}
    \hline
    \textbf{Population} & \textbf{Generations} & \textbf{Steel} & \textbf{Tungsten Alloy} & \textbf{Steel Dummy} & \textbf{inconel-718} \tabularnewline
    \hline
    50 & 50  & 0.9075 (1e-3)  & 0.8745 (2e-3)           & 0.9217 (1e-3)        & 0.8798 (2e-3)        \tabularnewline
    20 & 50  & 0.8907 (4e-3)  & 0.8572 (5e-3)           & 0.9073 (3e-3)        & 0.8583 (6e-3)        \tabularnewline
    50 & 100 & 0.9085 (1e-3)  & 0.8751 (1e-3)           & 0.9227 (9e-4)        & 0.8812 (1e-3)        \tabularnewline
    20 & 250 & 0.8926 (3e-3)  & 0.8582 (4e-3)           & 0.9103 (2e-3)        & 0.8598 (4e-3)        \tabularnewline
    \hline
\end{tabular}
\label{tab:reference-hypervolumes-pops-gens}
\end{table}

Table \ref{tab:CE-summary-pop-sizes} shows a summary of the worst, average, and best case for the computational costs obtained for search from scratch and adaption via NSGA-II, NSGA-II with varying goals, and NSGA-II with varying goals and active-inactive genotype, for the different setups of population size and maximum generations. One can observe that there is a sustained decrease in adaption cost for NSGA-II with varying goals and active-inactive genotype, as observed previously in section \ref{sec:results-costs} for population size 100. One exception is population size 20 with 50 generations, where the average and worst case costs are a bit higher in comparison to adaption with the baseline NSGA-II.

We present the results of a Friedman and post-hoc Nemenyi tests for each of the four groups in table \ref{tab:CE-summary-pop-sizes}, calculated with the CEs used to generate the summarized table. For all groups, the Friedman \textit{p-value} is lower than 0.01, which points to a difference between at least one pair of methods in each group. For population size 50 and 50 generations, the Nemenyi \textit{p-values} between the baseline NSGA-II and the three adaption variants are also incrementally significant (0.044 in comparison to baseline adaption, 0.006 to varying goals, and 0.001 to active-inactive genotype). For population size 20 and 50 generations, they are also significant, but higher for the varying-goals variant (0.001 in comparison to baseline adaption, 0.014 to varying-goals, and 0.001 to active-inactive genotype). For population size 50 and 100 generations, the \textit{p-value} is only significant between the baseline NSGA-II and NSGA-II with varying-goals and active-inactive genotype (0.002). For population size 20 and 250 generations, the baseline NSGA-II is different from the baseline adaption and active-inactive genotype (both \textit{p-values} of 0.001), but not from varying-goals. These results show that baseline adaption and NSGA-II with varying-goals may face some difficulties in certain setups, but the addition of the active-inactive genotype could overcome this.

\begin{table}
\center 
\caption{Worst, average, and best, cases for search from scratch and adaption when different population sizes are used, for baseline NSGA-II, NSGA-II with varying goals, and NSGA-II with varying goals and active-inactive genotype.}
\begin{tabular}{ccc|ccc}
    \hline
    \textbf{Algorithm} & \textbf{Popul.} & \textbf{Generat.} & \textbf{Worst Case} & \textbf{Average Case} & \textbf{Best Case} \tabularnewline
    \hline
    \textbf{From Scratch: Baseline}                    & 50 & 50 & 950 & 787 & 550 \tabularnewline
    \textbf{Adaption: Baseline}                        & 50 & 50 & 900 & 450 & 150 \tabularnewline
    \textbf{Adaption: Varying Goals}                   & 50 & 50 & 900 & 400 & 200 \tabularnewline
    \textbf{Adaption: Varying Goals + Active-Inactive} & 50 & 50 & 600 & 366 & 150 \tabularnewline
    \hline
    \textbf{From Scratch: Baseline}                    & 20 & 50 & 760 & 660 & 580 \tabularnewline
    \textbf{Adaption: Baseline}                        & 20 & 50 & 540 & 256 & 120 \tabularnewline
    \textbf{Adaption: Varying Goals}                   & 20 & 50 & 440 & 311 & 180 \tabularnewline
    \textbf{Adaption: Varying Goals + Active-Inactive} & 20 & 50 & 440 & 288 & 140 \tabularnewline
    \hline
    \textbf{From Scratch: Baseline}                    & 50 & 100 & 1,400 & 937 & 650 \tabularnewline
    \textbf{Adaption: Baseline}                        & 50 & 100 & 1,400 & 575 & 150 \tabularnewline
    \textbf{Adaption: Varying Goals}                   & 50 & 100 & 1,150 & 545 & 200 \tabularnewline
    \textbf{Adaption: Varying Goals + Active-Inactive} & 50 & 100 & 1,000 & 445 & 150 \tabularnewline
    \hline
    \textbf{From Scratch: Baseline}                    & 20 & 250 & 1,220 & 1,045 & 760 \tabularnewline
    \textbf{Adaption: Baseline}                        & 20 & 250 & 1,200 & 505   & 140 \tabularnewline
    \textbf{Adaption: Varying Goals}                   & 20 & 250 & 1,320 & 680   & 240 \tabularnewline
    \textbf{Adaption: Varying Goals + Active-Inactive} & 20 & 250 & 660   & 446   & 200 \tabularnewline
    \hline
\end{tabular}
\label{tab:CE-summary-pop-sizes}
\end{table}

Another observation is that, although the population sizes are lower, the costs for finding 99\% of the best hypervolume are of the same magnitude as with population size 100 (see table \ref{tab:CE-all-summary}). One possible reason is that, as the reduced population produces a Pareto front with a lower hypervolume, more generations are needed for optimization, even though the final hypervolume obtained is still worse than when using 100 individuals (see table \ref{tab:reference-hypervolumes}). Therefore, when more generations are available (100 and 250), some costs can be even higher. Concretely, the population size and number of generations used will depend on a trade-off between the cost of simulations and the desired quality for a specific application. In general, we can conclude based on our results with smaller populations that both the adaptive scheme and the extension with varying goals and active-inactive genotype are able to reduce the cost for finding a threshold solution with different population sizes, given that a population was already trained on a source task.

\section{Conclusion and Future Work}
\label{sec:conclusion}

We have addressed the issue of reducing the number of computations necessary for manufacturing process optimization, which usually requires costly simulations, in the context of changing product specifications. For that purpose, we have considered optimization algorithms from the viewpoint of system flexibility, which is related to dynamic optimization. We have studied the ability of an optimization algorithm to adapt a solution for a target task that has previously been found for a source task, thus reducing the cost for finding a new solution.

In order to be able to systematically experiment within this framework, we have used the extended Oxley model, which simulates the process of orthogonal metal cutting. We introduced a new benchmark in the form of a multi-objective problem based on the extended Oxley model. We used the NSGA-II algorithm to optimize the process parameters for different tasks defined by different material parameters, and experimented with adapting solutions for pairs of source and target tasks, in order to assess the potential of adaption in comparison to search from scratch. Additionally, we extended NSGA-II with two features inspired by facilitated variation and that relate to dynamic optimization: varying goals, where a population is evolved for two goals at the same time, and active-inactive genotype, where each gene contains different possible values and only one is active at a time. 

Given that the model used for our proposed benchmark problem is simpler than a numerical simulation, our results are a starting point and can be further extended by considering extensions to this model. However, it already enabled us to perform a more comprehensive analysis of the proposed methods and to show that they are able to reduce the adaption cost in a context of manufacturing optimization. Specifically, we could draw the following main conclusions:

\begin{enumerate}
    \item When problems are related and one can expect that solutions lie in a similar region of the search space, adapting solutions from source to target greatly reduces the number of evaluations needed for finding a threshold solution. The extensions with the varying goals strategy and active-inactive genotype can further reduce this cost by generating solutions that are in-between original solutions and accommodating different possibilities/structures in one genotype.
    
    \item If a lower population size is enough for finding a desired solution for a given problem, the adaptive scheme and proposed variants most likely will still provide an advantage over search from scratch, given that a solution for one or more related tasks was already found.
\end{enumerate}

As outlined in the introduction, the introduced benchmark allows only for negative conclusions.
Therefore, the benchmark itself currently should be considered as a proof-of-concept. Its usefulness for identifying flexible optimization methods for manufacturing is only proven once an optimization method developed on the basis of the benchmark has been positively validated with more realistic simulations.

Some interesting possibilities of future work that could follow from the work presented in this paper are:

\begin{enumerate}
    \item A more concrete definition of the multi-objective problem based on the Oxley model. Although we have used the hypervolume as a measure of the quality of the Pareto fronts, a more concrete definition of a goal would make the gains a given algorithm can bring more understandable in practice. Furthermore, the Oxley model could be extended to reflect more the properties of real simulations, which could make the defined problem as a candidate benchmark for manufacturing optimization.
   
    \item The defined multi-objective problem enables fast evaluation of different algorithm setups. A natural next step is the validation of the best setups found in a scenario based on, for example, FE simulations.
   
    \item Although the presented results show an improvement in the cost for adaption, the values obtained are still high if considered in a practical setting. Given the interdisciplinary flexibility framework used, one could consider extending the ideas of this work with based on ideas for adapting and transferring data-driven models from other areas, such as meta-learning~\cite{vanschoren2019meta}, transfer learning~\cite{yang2020transfer}, few-shot learning~\cite{wang2020generalizing}, as well as efforts to combine transfer learning and dynamic multi-objective optimization~\cite{ruan2019and}.
    
    \item The proposed extensions to NSGA-II can also be incorporated in other optimizers that are based on iterations. One interesting possibility is integrating the varying goals and active-inactive genotype strategy into a Bayesian optimizer, which is a popular method for manufacturing optimization.
    
    %\item A less direct but still interesting line of study is the understanding of the mechanisms behind the proposed strategies, like, for example, what happens during different epochs, what percentage of constructive mutations are associated with activations and/or deactivations.
\end{enumerate}

%\begin{acks}
%To Robert, for the bagels and explaining CMYK and color spaces.
%\end{acks}

%\newpage

\bibliographystyle{ACM-Reference-Format}
\bibliography{references}

%\appendix


\end{document}